\documentclass[letterpaper]{article} 
\usepackage{aaai2026}  
\usepackage{times}  
\usepackage{helvet}  
\usepackage{courier}  
\usepackage[hyphens]{url}  
\usepackage{graphicx} 
\usepackage{natbib}  
\usepackage{caption} 
\usepackage{algorithm}
\usepackage{algorithmic}
\usepackage{booktabs}   
\usepackage{multirow}   
\usepackage{makecell}   
\usepackage{colortbl}   
\usepackage{xcolor}     
\usepackage{graphicx} 
\usepackage{amsmath}    
\usepackage{amssymb}    
\usepackage{bm}         
\usepackage{newfloat}
\usepackage{listings}

\usepackage{newfloat}
\usepackage{listings}
\usepackage{enumitem}

\usepackage{algorithm}
\usepackage{algorithmic}
\usepackage{booktabs} 
\usepackage{multirow}
\usepackage{amsmath}
\usepackage{algorithmic}
\usepackage{cleveref}
\usepackage{booktabs}
\usepackage{xcolor}

\definecolor{Ocean}{RGB}{129,194,234}

\DeclareCaptionStyle{ruled}{labelfont=normalfont,labelsep=colon,strut=off} 
\floatstyle{ruled}
\newfloat{listing}{tb}{lst}{}
\floatname{listing}{Listing}
\title{MobileRAG: Enhancing Mobile Agent with Retrieval-Augmented Generation}
\author{
    Gowen Loo\textsuperscript{\rm 1}\equalcontrib, Chang Liu\textsuperscript{\rm 2}\equalcontrib, Qinghong Yin\textsuperscript{\rm 6}, Xiang Chen\textsuperscript{\rm 4}, Jiawei Chen\textsuperscript{\rm 5}, Jingyuan Zhang\textsuperscript{\rm 6}, Yu Tian\textsuperscript{\rm 7}\thanks{Corresponding author}
}
\affiliations{

    \textsuperscript{\rm 1} Network and Data Security Key Laboratory of Sichuan Province, University of Electronic Science and Technology of China\\
    
    \textsuperscript{\rm 2}DEPARTMENT OF COMPUTING, Hong Kong Polytechnic University\\

    \textsuperscript{\rm 4}College of Computer Science and Technology, Nanjing University of Aeronautics and Astronautics\\

    \textsuperscript{\rm 5}East China Normal University \textsuperscript{\rm 6}Kuaishou-Inc\\
    \textsuperscript{\rm 7}Dept. of Comp. Sci. and Tech., Institute for AI, Tsinghua University, Tsinghua University
    
    202314090108@std.uestc.edu.cn, chang181.liu@connect.polyu.hk,
    tianyu181@mails.ucas.ac.cn.
}

\usepackage{bibentry}

\begin{document}

\maketitle

\begin{abstract}
Smartphones have become indispensable in people's daily lives, permeating nearly every aspect of modern society. With the continuous advancement of large language models (LLMs), numerous LLM-based mobile agents have emerged. These agents are capable of accurately parsing diverse user queries and automatically assisting users in completing complex or repetitive operations. However, current agents 1) heavily rely on the comprehension ability of LLMs, which can lead to errors caused by misoperations or omitted steps during tasks, 2) lack interaction with the external environment, often terminating tasks when an app cannot fulfill user queries, and 3) lack memory capabilities, requiring each instruction to reconstruct the interface and being unable to learn from and correct previous mistakes. To alleviate the above issues, we propose MobileRAG, a mobile agents framework enhanced by Retrieval-Augmented Generation (RAG), which includes InterRAG, LocalRAG, and MemRAG. It leverages RAG to more quickly and accurately identify user queries and accomplish complex and long-sequence mobile tasks. Additionally, to more comprehensively assess the performance of MobileRAG, we introduce MobileRAG-Eval, a more challenging benchmark characterized by numerous complex, real-world mobile tasks that require external knowledge assistance. Extensive experimental results on MobileRAG-Eval demonstrate that MobileRAG can easily handle real-world mobile tasks, achieving 10.3\% improvement over state-of-the-art methods with fewer operational steps. Our code is publicly available at: \url{https://github.com/liuxiaojieOutOfWorld/MobileRAG_arxiv}
\end{abstract}

\section{Introduction}
Today, approximately 4.69 billion individuals own smartphones, with average daily online activity on smartphone exceeds 3 hours and 30 minutes\footnote{\url{https://backlinko.com/smartphone-usage-statistics}}. Users increasingly rely on smartphones to handle various tedious and complex tasks. With the rapid development of large language models (LLMs)~\cite{hurst2024gpt,team2024gemini,guo2025deepseek} , numerous LLM-based mobile agents~\cite{wang2024mobile,wang2024mobilev2,wang2025mobile,wu2024foundations} have emerged. These agents can accurately parse users' diverse instructions and assist users automatically complete complex or repetitive operations. As shown in Figure \ref{fig:visual}, despite these advancements, three key challenges remain in practical applications:

\begin{figure}[!t]
    \centering
    \includegraphics[width=0.9\linewidth]{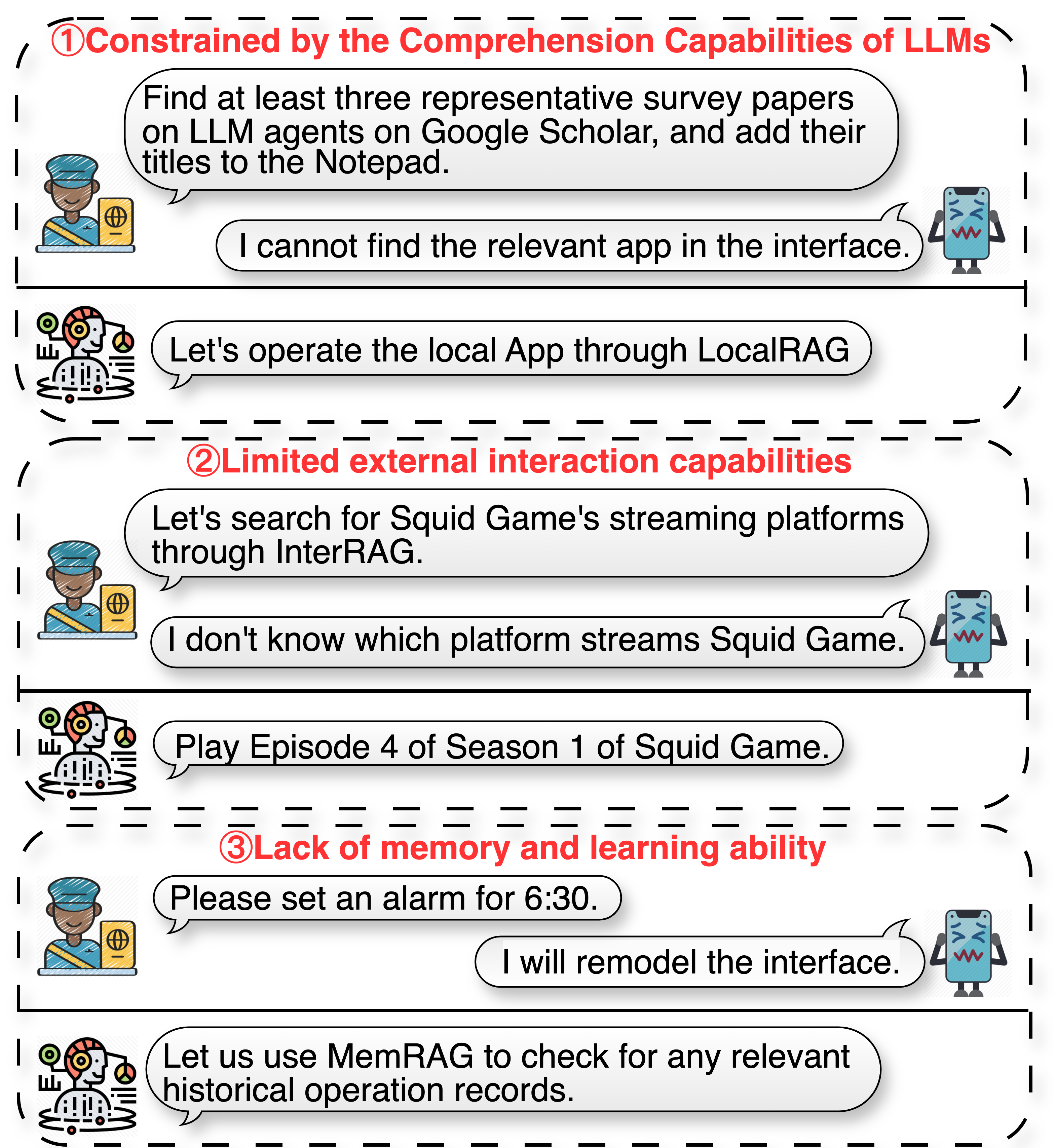}
    \caption{Three key challenges for mobile agents in practical applications.}
    \label{fig:visual}
\end{figure}

1) Previous works commonly rely excessively on LLMs’ comprehension capabilities~\cite{wen2023empowering,wang2024mobile,wen2024autodroid}, necessitating multiple sequential steps for even basic mobile device operations. The absence of robust reasoning mechanisms leads to significant performance deficiencies, including operational errors and step omissions when handling complex tasks. Even seemingly simple application interactions require multiple cognitive processes (such as interface comprehension, apps localization, and execution initiation). This fundamental limitation undermines the agent's autonomous task completion abilities and substantially constrains its practical utility in real-world deployment scenarios.

2) Current agents~\cite{lee2024mobilegpt, zhang2025appagent,wang2025mobile} generally exhibit limited capability to interact effectively with external environments. confronted with application functionality constraints or unforeseen circumstances during task execution, these agents frequently opt to terminate the task rather than explore alternative solution pathways. For instance, when a user inputs "I want to watch Squid Game," existing mobile agents often abandon the task due to insufficient knowledge of available streaming platforms. This inflexibility hampers the agents’ adaptability and makes them poorly equipped to handle complex, dynamic real-world environments.

3) Mobile agents demonstrate  pronounced deficiencies in memory and learning capabilities. Most existing works are incapable of accumulating experience from prior tasks, necessitating the reconstruction of operational interfaces with each new instruction. These approaches not only result in inefficient use of computational resources, but also prevent the system from learning from historical successful cases and engaging in self-correction. Consequently, these limitations undermine both the operational efficiency of agents and their capacity for autonomous, long-term evolution.

Retrieval-Augmented Generation (RAG) introduces a novel research direction for enhancing LLMs’ comprehension~\cite{lewis2020retrieval,gao2023retrieval,fan2024survey}, external interaction, and memory learning capabilities. However, the direct application of RAG to mobile agents presents several challenges: 1) agents may struggle to discern when to invoke internal versus external applications; 2) limitations imposed by the need for model lightweighting; and 3) inherent trade-offs in memory retention mechanisms. To address these issues, we propose MobileRAG, a RAG-based framework enhanced for mobile agents. This framework comprises three core components—InterRAG, LocalRAG, and MemRAG—that collectively aim to leverage RAG technologies for more rapid and accurate identification of user intents and for efficient execution of complex, long-sequence device operations. Specifically, InterRAG facilitates the invocation of external applications and supports robust information retrieval, thereby enhancing the agent’s capacity for external interaction; LocalRAG streamlines on-device deployment and inference, constructing a lightweight RAG model optimized for mobile environments; and MemRAG tackles the challenge of memory retention by preserving representative and successful user actions, thus improving the real-time responsiveness and overall efficiency of MobileRAG.

To verify the effectiveness of MobileRAG and its capability to interact with external knowledge, we construct MobileRAG-Eval, a more challenging benchmark designed to require deeper semantic understanding and integration of external knowledge. Extensive experimental results on MobileRAG-Eval demonstrate that MobileRAG achieves superior performance in leveraging external knowledge and managing complex, long-sequence tasks, with a 10.3\% improvement over state-of-the-art (SOTA)~\cite{wang2025mobile} methods. Notably, our framework significantly streamlines the process and reduces the likelihood of erroneous operations by directly retrieving apps via RAG. In summary, our main contributions are as follows:
\begin{itemize}
	\item We design a RAG-based framework enhanced for mobile agents, comprising InterRAG, LocalRAG, and MemRAG. It leverages RAG technology to more quickly and accurately identify customer needs and accomplish more complex and long-sequence device operation tasks.
	\item Through the effective interaction of InterRAG and LocalRAG, MobileRAG enables agents to incorporate external knowledge, facilitating a more comprehensive understanding of user queries and supporting accurate, efficient operation of mobile devices.
	\item With the memory capabilities of MobileRAG, successful cases from previous operations can be retained and user instructions can be matched to historical instances using MemRAG. It further reduces unnecessary analysis and operational steps, improving both speed and accuracy.
\end{itemize}
\section{Related Work}

\subsection{Mobile Agents}

In recent years, the rapid development of LLMs has driven significant advances in mobile agents. Broadly, existing mobile agents can be divided into two main categories. The first category consists of XML-based mobile agents, which utilize XML~\cite{cabri2000xml,ciancarini2002coordination,tamayo2011instance} as a medium for structured data exchange. Owing to the universality and high readability of XML, these agents~\cite{chen2008xml,chou2010embeddable} facilitate information integration and collaboration across heterogeneous systems. However, limited by the data representation capabilities and processing mechanisms of XML, such agents often lack the flexibility needed to adapt to complex and dynamic task requirements. The second category comprises GUI-based agents~\cite{sun2022meta,li2022spotlight,gou2024navigating}, which emphasize interaction through visual interfaces~\cite{hong2024cogagent} to enable real-time decision-making and task execution. Nevertheless, current GUI-based agents are entirely reliant on LLMs for interface understanding~\cite{wen2023empowering,rawles2023androidinthewild,gur2023real}. This dependency can lead to issues with model recognition and decision accuracy in real-world applications, potentially impacting the reliable completion of tasks. To address these issues, we propose MobileRAG, a framework designed to accurately understand user queries, effectively perform information retrieval, and enhance memory and learning capabilities.

\begin{figure*}[!t]
    \centering
    \includegraphics[width=0.9\linewidth]{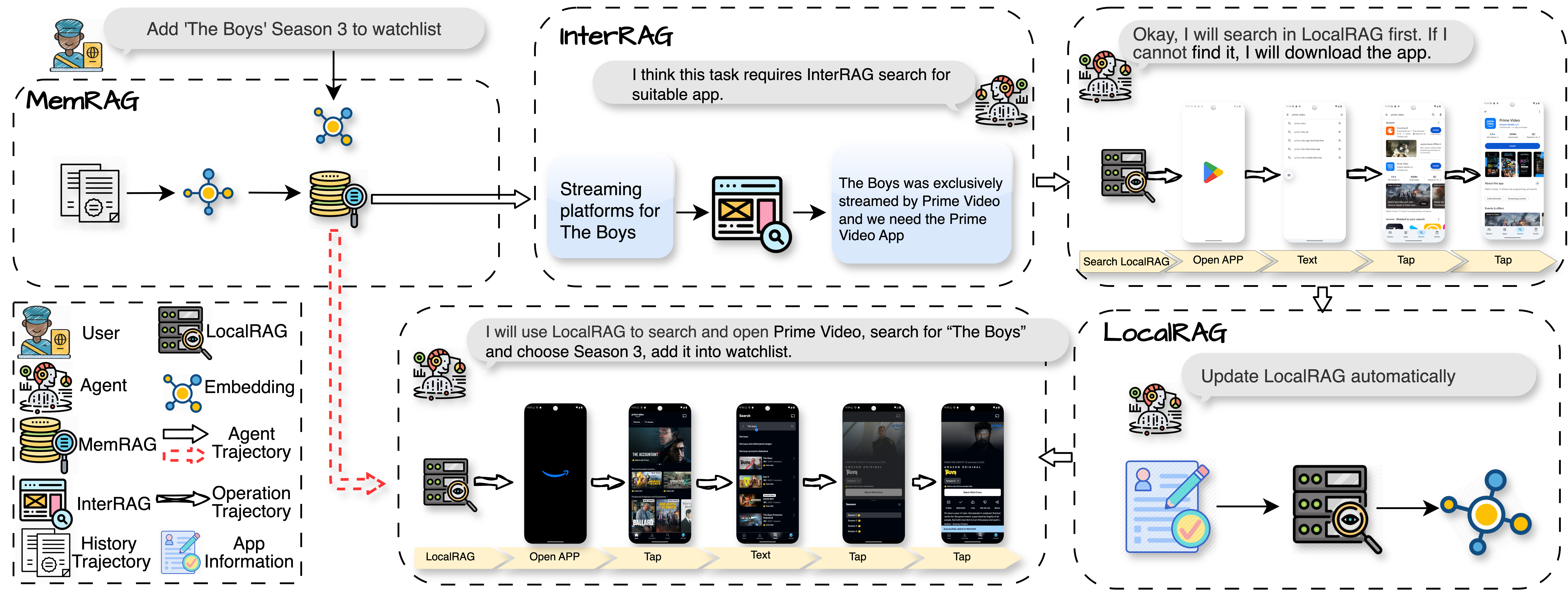}
    \caption{Overall framework of MobileRAG. It comprises three key RAG modules: 1) LocalRAG, responsible for managing and retrieving local applications; 2) InterRAG, dedicated to retrieving external knowledge; and 3) MemRAG, which manages and retrieves similar successful historical steps. The \textcolor{red}{red paths} are exclusively available in the retrieved historical steps.}
    \label{framework}
\end{figure*}

\subsection{Retrieval-Augmented Generation}

RAG is an emerging hybrid architecture that addresses the limitations of purely generative models ~\cite{zhao2024retrieval,chen2024benchmarking,es2024ragas}. It improves answer accuracy by retrieving relevant information and generating responses based on that data. While RAG has been widely applied in fields such as LLMs ~\cite{gao2023retrieval,fan2024survey}, LVLMs  ~\cite{yu2024visrag}, and agent-based systems  ~\cite{lala2023paperqa}, its adopt in mobile agents remains underexplored ~\cite{gupta2024comprehensive}. The key challenges here include difficulties in accurately determining when to invoke internal or external applications, constraints imposed by the limited computational power and storage capacity of mobile devices, and trade-offs inherent in memory retention mechanisms between maintaining information integrity and ensuring processing efficiency. Collectively, these issues hinder task execution efficiency and ultimately degrade the overall performance of mobile agents. To address these challenges, we propose MobileRAG, a RAG-based framework designed for mobile agents. It includes three key components: InterRAG, which enables interaction with external applications for up-to-date information; LocalRAG, which handles on-device knowledge retrieval for faster access; and MemRAG, which uses memory-based learning to enhance accuracy and task completion by remembering successful operations. These components collectively address existing limitations, improve execution efficiency, and boost mobile agent performance.

\section{Methods}

\subsection{Overview of MobileRAG}

To address inadequate reasoning, limited adaptability to external interactions, and poor memory capabilities of existing mobile agents, we propose MobileRAG, a RAG-based framework specifically designed for mobile environments. While RAG has shown promise in various domains, directly applying it to mobile scenarios introduces several challenges, including distinguishing between internal and external resources, managing lightweight model constraints, and balancing memory retention trade-offs. MobileRAG resolves these issues by integrating tailored components to enhance mobile agent performance and efficiency.

MobileRAG consists of three components: MemRAG, LocalRAG, InterRAG. As shown in Figure \ref{framework}, upon receiving a user query, it first utilizes MemRAG to determine whether identical or similar tasks have previously been executed successfully. For exact matches, the system directly reuses the recorded operation steps; for similar queries, these prior cases are provided to the agent as guidance. During task execution, when the agent encounters unfamiliar knowledge or requires  additional context, it dynamically retrieves external information via InterRAG. Subsequently, LocalRAG is responsible for verifying whether the necessary apps are already installed on the device, thereby supporting efficient and accurate app management. If an app is not available locally, LocalRAG initiates its download from Play Store and updates the local application database accordingly. After confirming the availability of all required resources and apps, the agent integrates these components, executing precise steps including app launching, navigation, content search, and task completion. Through the coordinated collaboration of MemRAG, InterRAG, LocalRAG, MobileRAG seamlessly transforms complex user queries into actionable and automated workflows on mobile devices.

\subsection{LocalRAG}

To mitigate operational errors or omitted steps caused by mobile agents' complete reliance on agent's comprehension capabilities, we propose LocalRAG, specifically designed to directly perceive and manage locally installed app. We select the BGE-small series as our foundational retrieval model to address the limitations associated with lightweight retrieval architectures. To achieve robust and efficient semantic matching, We design an optimized retrieval process tailored to local apps, enabling effective matching 
between user prompts and apps while promptly identifying cases where no suitable match exists. Specifically, we employ a LLM to generate user query, for interacting with the local app based on its description \footnote{we collected the ``About this app'' section from Google Play as the app description.}. To enable LocalRAG to identify cases where no suitable app exists, we generate a set of user query  and the “None” indicator bit that are not supported by any local apps. By performing supervised contrastive learning on the retriever between the user query and the app descriptions, we ensure accurate alignment between user task expressions and app functionalities in mobile scenarios, while also equipping the model with the capability to effectively reject unsupported commands.

Upon receiving a user query, LocalRAG calculates the cosine similarity between the query and the app descriptions within the local app library. It subsequently returns the top-3 relevant apps informations to the agent, providing comprehensive metadata including app names, package identifiers, descriptions, and similarity scores. The agent determines the appropriate app by analyzing the user's query, and then triggers the selected app through its package identifier. When LocalRAG yields no matches or the highest similarity score falls below the established threshold, it notifies the agent that no appropriate local app is available, prompting LocalRAG to initiate the downloading process from Play Store. Subsequently updates its local app database accordingly.

\subsection{InterRAG}

Previous work lacks adaptive mechanisms to confront with unknown contexts or app constraints, thus tend to terminate tasks rather than explore possible alternatives. This restricts their operation to the boundaries of predefined knowledge and capabilities. To address these limitations and empower mobile agent to dynamically extend its knowledge boundaries, we introduce InterRAG, a module designed to enhance agent capabilities through real-time access to external information sources. It systematically augments agent in comprehending and processing user queries that involve previously unknown contexts, thereby facilitating continuous knowledge acquisition and adaptive response generation.

As a component for interacting with external knowledge, InterRAG significantly enhances the agent's adaptability to open-domain tasks. When the agent encounters entities, terms, or concepts beyond its training corpus or local app base during user query processing, it delegates dynamic retrieval to InterRAG. Specifically, InterRAG formulates queries for unfamiliar keywords or entities and acquires real-time information from the internet via the Google Search API. It parses and filters the search results, extracting the top-10 relevant results and returning them to the LLM in a structured format that includes titles and summaries. Leveraging this up-to-date external information, the agent can more accurately interpret the semantic intent of user queries and recommend suitable apps or actions accordingly. Notably, when apps within the local app base prove insufficient for user requirements, InterRAG enables the agent to download relevant apps from Google Play and incorporate their information into the local app base, thus optimizing subsequent retrieval efficiency. This workflow allows the mobile agent to flexibly respond to queries involving newly emerging or rapidly evolving entities, substantially broadening its operational scope and adaptability in open-domain tasks.

\subsection{MemRAG}

The existing mobile agent frameworks lack the ability to leverage successful experiences from previous tasks to enhance its operational capabilities. Upon receiving new queries, the agent must reconstruct the operational interface from scratch, leading to inefficient utilization of computational resources and an inability to self-correct based on prior successes. To address these limitations and strengthen the agent’s memory and learning capabilities, we propose MemRAG, a module designed to efficiently store and manage records of successful operations. During the processing of new tasks, MemRAG enables the agent to automatically retrieve and reference relevant historical information, facilitating operation reusability and promoting continuous self-improvement of system capabilities.

Upon successful execution of a user’s query, MobileRAG stores the user’s query alongside the corresponding operation steps in the memory database. When a new query is received, It employs MemRAG to retrieve relevant historical queries. If the similarity between the most relevant historical query and the current query exceeds a predefined threshold, the corresponding historical operation step is provided to the agent alongside the new query to inform the generation of an updated execution path. In cases where the queries are determined to be identical, the system bypasses step regeneration and executes the task directly using the established operational sequence. MemRAG fundamentally enhances the agent with sophisticated memory and learning capabilities. As user queries with the device through the agent accumulate, the mobile agent undergoes continuous evolution, progressively strengthening its autonomous capabilities. Additionally, MemRAG improves operational efficiency through two key advantages: 1) fewer operation steps accelerate the agent’s response speed, 2) a reduced set of steps minimizes the influence of erroneous cases, thereby improving the overall accuracy of the agent’s outputs.

\section{Experiments}

\begin{figure*}[!t]
    \centering
    \includegraphics[width=0.9\linewidth]{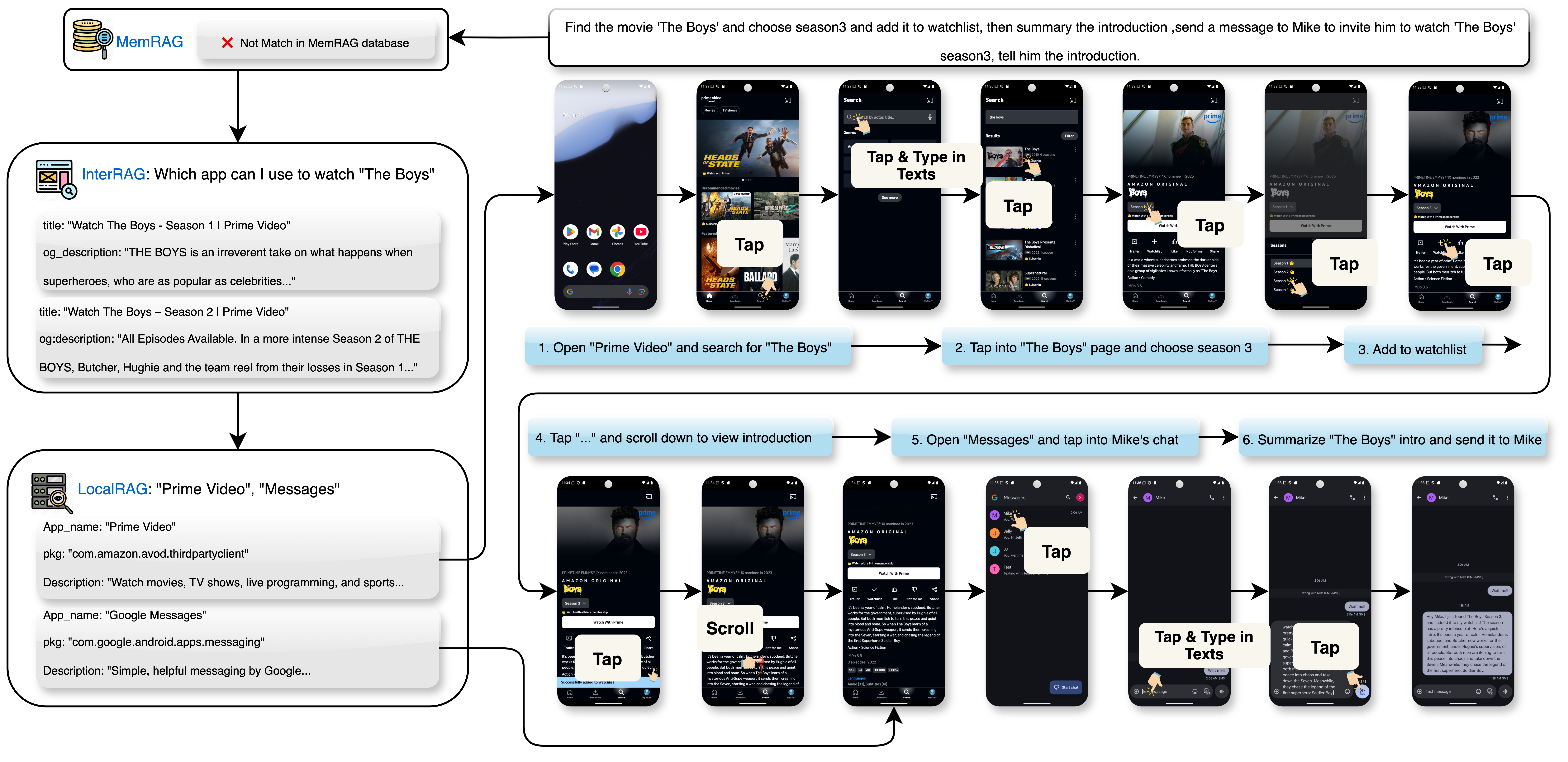}
    \caption{Leveraging LocalRAG and InterRAG for complex user query workflows. InterRAG effectively utilizes external knowledge to interpret user queries, while LocalRAG reduces operational errors or omitted steps that may arise from the LLMs’ insufficient understanding of interfaces, thereby enabling the modeling of long sequential tasks.}
    \label{exp1}
\end{figure*}

\begin{table}[t]

\begin{center}
\begin{small}

\centering
\setlength{\tabcolsep}{1mm} 
\begin{tabular}{lccccccc}
\toprule
Benchmark &
Tasks & 
\begin{tabular}[c]{@{}c@{}} Multi-App\\Tasks \end{tabular} & 
\begin{tabular}[c]{@{}c@{}} No-App\\Tasks \end{tabular} & Apps & 
\begin{tabular}[c]{@{}c@{}} Avg \\Ops \end{tabular} & 
\begin{tabular}[c]{@{}c@{}} Total \\Ops \end{tabular} \\
\midrule
Mobile-Eval & 33 & 3 & 9 & 10 & 5.55 & 183 \\
Mobile-Eval-v2 & 44 & 4 & 8 & 10 & 5.57 & 245 \\
AppAgent & 45 & 0 & 0 & 9 & 6.31 & 284 \\
\midrule
Our Dataset & 80 & 20 & 30 & 20 & 8.01 & 641 \\
\bottomrule
\end{tabular}
\end{small}
\end{center}
\caption{Comparison with existing dynamic evaluation benchmarks on emulator devices.}
\label{tab:benchmark}
\end{table}

\subsection{MobileRAG-Eval}
As shown in Table \ref{tab:benchmark}, existing benchmarks predominantly focus straightforward tasks with predefined app specifications, where performance has largely plateaued. However, they remain constrained by static local apps, failing to accommodate dynamic app updates, real-time external information integration, and capabilities for longitudinal task reasoning and operational learning. To address these limitations, we introduce MobileRAG-Eval, a challenging benchmark designed for dynamically updating on-device apps that mandates external knowledge augmentation in complex real-world scenarios, emphasizing real-time constraints, reasoning-intensive planning, and cross-app coordination. It evaluates agents to autonomously select optimal tools, intelligently interface with external services to overcome task execution barriers, and self-correct based on historical demand patterns and operational feedback. We detail the benchmark specifics in Appendix \textcolor{red}{A}.

MobileRAG-Eval integrates the MobileAgent-v2 benchmark~\cite{wang2024mobilev2} and incorporates 36 novel advanced instructions, significantly extending both app coverage and task complexity. Its core innovation lies in the establishment of a progressive three-tier evaluation framework that spans atomic operations, composite tasks, and open-environment instructions. Specifically, the framework includes: (1) a basic atomic operation library (core interaction primitives within standalone apps); (2) multi-app coordination directives (chained cross-app procedures); and (3) open-scenario instructions (application-agnostic tasks requiring external knowledge retrieval and autonomous tool selection). The expanded dataset encompasses a dynamically evolving app ecosystem, covering three critical domains: core device functionalities (system configuration, app lifecycle management), high-frequency lifestyle services (multimedia resource access, geolocation navigation, audio control), and cross-app workflows. This comprehensive structure enables a systematic assessment of agents' capabilities in adaptive tool orchestration, real-time decision-making robustness, and disturbance resistance within dynamic, uncertain environments. 

\subsection{Experimental Setting}

\textbf{Metrics}. We propose a comprehensive five-dimensional evaluation suite to quantify agent capabilities in dynamic mobile environments. The metrics are detailed as follow:
\begin{itemize}[leftmargin=0.2cm]
    \item \textbf{App Selection (AS)} is defined as the percentage of tasks in which the optimal app is selected, calculated as: AS = Correct app selections / Total app selections.

    \item \textbf{Action Fidelity (AF)} measures the proportion of correctly executed atomic actions (Tap, Type, Swipe, Back, Stop) relative to ground truth sequences, calculated as AF = Correct actions / Total actions.
 
    \item \textbf{Reflection Precision (RP)} is the correct diagnosis rate during error recovery, calculated as AF = Correct reflections / Total reflections.

    \item \textbf{Task Completion Ratio (TCR)} measures the proportion of achieved sub-goals with atomic verification, calculated as TCR = Completed sub-goals / Total sub-goals.

    \item \textbf{Task Success Rate (TSR)} is defined as Binary success measurement where success requires all user query  requirements fulfilled. calculated as TSR = Successful tasks / Total tasks.
\end{itemize}

\begin{table*}[t]
\begin{center}
\begin{small}
\setlength{\tabcolsep}{1.6 mm}
\centering
\begin{tabular}{lc|ccccc}
\toprule
Model & 
Type &
\begin{tabular}[c]{@{}c@{}} App Selection \\(\%)  $\uparrow$ \end{tabular} & 
\begin{tabular}[c]{@{}c@{}} Action Fidelity\\(\%)  $\uparrow$ \end{tabular} & 
\begin{tabular}[c]{@{}c@{}} Reflection Precision\\(\%) $\uparrow$ \end{tabular} & 
\begin{tabular}[c]{@{}c@{}} Task Completion Ratio\\(\%) $\uparrow$ \end{tabular} & 
\begin{tabular}[c]{@{}c@{}} Task Success Rate\\(\%) $\uparrow$ \end{tabular} \\
\midrule
Mobile-Agent-v1& Single-Agent & 76.3 & 65.8 & - & 63.1 & 47.5 \\
Mobile-Agent-v2 & Multi-Agent & 77.5 & 71.2 & 95.8 & 66.3 & 52.5 \\
Mobile-Agent-E & Multi-Agent & 92.9 & 87.6 & 98.8 & 88.6 & 72.5\\
\midrule
MobileRAG & Multi-Agent & 100.0 & 86.4 & 98.5 & 91.2 & 80.0 \\
\midrule
MobileRAG$^*$ & Multi-Agent &  100.0 & 87.8  & 98.8  &  92.9 & 82.5  \\
\bottomrule
\end{tabular}

\end{small}
\end{center}
\caption{Comparison with SOTA on the MobileRAG-Eval. $^*$ indicates results obtained on the Xiaomi 15 Pro, all other results are collected on the Pixel 9 within the Android Studio environment.}
\label{tab:main_result_gpt4o}
\end{table*}

\noindent\textbf{Baseline}. We comprehensively evaluate our method against three state-of-the-art (SOTA) open-source mobile agent frameworks, including Mobile-Agent~\cite{wang2024mobile}, Mobile-Agent-v2~\cite{wang2024mobilev2} , and Mobile-Agent-E~\cite{wang2025mobile}. For fair comparisons, We compared MobileRAG with previous work on Pixel 9 in Android Studio, all baselines are implemented under identical experimental conditions, sharing the same atomic action space, visual perception models, and initial prompting strategies as MobileRAG. To further validate the framework’s effectiveness in real-world scenarios, we performed extensive experiments on the Xiaomi 15 Pro.

\begin{table}[htbp]

		\begin{tabular}{l|ccccc}
			\toprule
			MobileRAG &   AS & AF & RP & TCR & TSR\\ 
			\midrule
			Gemini-1.5-pro     & 96.3 & 74.3 & 97.2 & 90.2 & 71.3\\
			Claude-3.5-Sonnet    & 100.0 & 87.1 & 98.7 & 92.7 & 83.8 \\
                GPT-4o    & 100.0 &86.4 & 98.5&    
            91.2 &  80.0\\
            \bottomrule
		\end{tabular}

		\caption{Results on various LLMs backbones}
		\label{LLMs}	
\end{table}

\noindent\textbf{Implementation Details}. Our agent consists of two primary stages: the App Selection stage and the Action stage, each leveraging advanced LLMs as backbone, including Claude-3.5-Sonnet~\cite{anthropic2024model}\footnote{Claude-3.5 version: Claude-3.5-Sonnet-2024-10-22}, Gemini-1.5-pro~\cite{team2024gemini}\footnote{Gemini-1.5 version: Gemini-1.5-Pro-2024-04-09}, and GPT-4o~\cite{hurst2024gpt}\footnote{GPT-4o version: GPT-4o-2024-11-20}. Unless specified otherwise, GPT-4o serves as the default backbone for all models. In the App Selection stage, we adopt BGE-small as retrieval model to enhanced context retrieval and accurate app selection. In the Action stage, the agent integrates a comprehensive Visual Perception Module composed of the ConvNextViT-document OCR model~\cite{liao2020real} from ModelScope for text recognition, GroundingDINO~\cite{liu2024grounding} for natural language prompt-based object detection, and Qwen-VL-Int4~\cite{bai2023qwen} for generating detailed icon descriptions, collectively enabling precise visual interaction within mobile environments. A similarity threshold of 0.8 is adopted in MemRAG, ensuring that only closely related historical queries inform the response to new inputs.

\subsection{Experiment Results}

\textbf{Comparison with SOTA frameworks.} Table \ref{tab:main_result_gpt4o} presents the result of MobileRAG and all baselines on MobileRAG-Eval. The results demonstrate that our method achieves superior perfor mance on various metrics, especially on AS, MobileRAG can achieve a 100\% success rate, confirming the effectiveness of LocalRAG. By constructing app information into a local knowledge base, it can effectively mitigate operational errors or omitted steps caused by mobile agents’ complete reliance on LLM’s comprehension capabilities. Meanwhile, we achieve SOTA performance on TSR, achieveing 10.3\% improvement over other baseline. This result demonstrates that MobileRAG can effectively interpret users’ ambiguous intentions via InterRAG, making it particularly well-suited for real-world application scenarios.

\begin{table}[htbp!]
    \setlength{\tabcolsep}{1 mm}
    \small
    \begin{tabular}{l|rrr|rrr}
    \toprule
    \multicolumn{1}{c|}{\multirow{2}[2]{*}{Model}} & \multicolumn{3}{c|}{LLM} & \multicolumn{3}{c}{Action} \\
    \multicolumn{1}{c|}{} & \multicolumn{1}{l}{Single} & \multicolumn{1}{l}{Dual} & \multicolumn{1}{l|}{Triple} & \multicolumn{1}{l}{Single} & \multicolumn{1}{l}{Dual} & \multicolumn{1}{l}{Triple} \\
    \midrule
    Mobile-Agent-v2    &   3.0    &   7.0    &    11.3   &   3.0    &   7.0    &  11.3\\
    Mobile-Agent-E     &    1.9   &   4.8    &    8.0   &  1.9     &   4.8    &  8.0\\
    \midrule
    MobileRAG  &       1.9    &    3.7   &   4.3    &    1.0   & 1.0 & 1.0  \\
    \bottomrule
    \end{tabular}%
  
  \caption{Efficiency of MobileRAG on Single-, Dual-, and Triple-App Tasks. ``LLM'' indicates the number of LLM calls, and ``Action'' indicates the number of mobile steps.}
  \label{efficiency}%
\end{table}%

We further validated the effectiveness of MobileAgent in real-world environments (Xiaomi 15 Pro). As shown in Table \ref{tab:main_result_gpt4o}, MobileAgent performs exceptionally well across several key indicators and can reliably handle complex and dynamic practical app scenarios. Notably, compared to XML-based and GUI-based mobile agents, MobileRAG can directly retrieve relevant context via LocalRAG, efficiently access and launch target apps, and flexibly leverage system-level functionalities. Owing to its lightweight architecture and high compatibility, it is well-suited for widespread deployment on real-world consumer mobile devices.

Table \ref{LLMs} presents a comparison of MobileRAG across different backbone LLMs. MobileRAG consistently outperforms across all recent LLMs, owing to its superior framework design. It effectively leverages retrieval of historical information, local apps, and web data, which substantially mitigates comprehension or operational errors caused by large model hallucinations, and greatly enhances the reliability and practical value of the system.

\begin{figure*}[!t]
    \centering
    \includegraphics[width=0.92\linewidth]{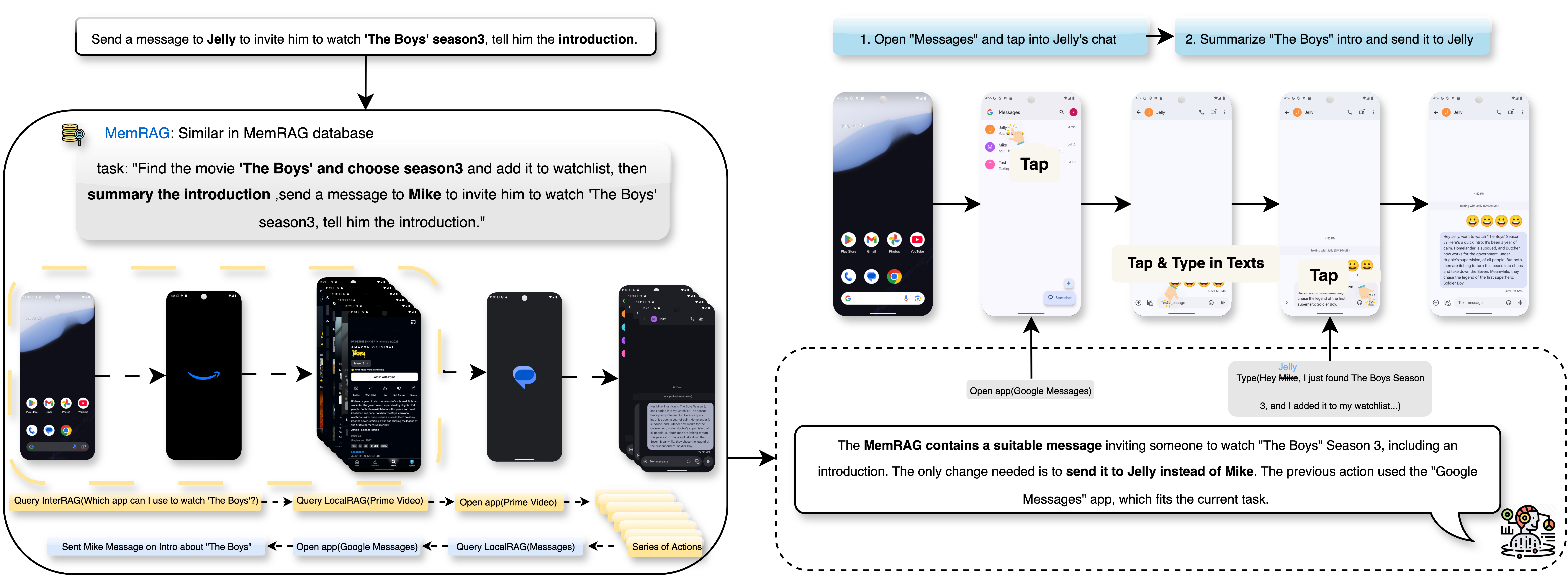}
    \caption{Leveraging MemRAG to retrieve previously successful steps enables the mobile agent to perform rapid operations. With its advanced memory and learning mechanisms, MemRAG empowers the agent to adapt based on past successes. As users interact with the agent and more queries accumulate, the mobile agent evolves, gradually enhancing its autonomous capabilities.}
    \label{case_mem}
\end{figure*}

\textbf{Efficiency Analysis.} Table \ref{efficiency} reports the efficiency of MobileRAG by quantifying the number of steps required for LLM-based agents to invoke and open apps, an essential factor in evaluating the model’s practical applicability. Compared to other methods, MobileRAG demonstrates a substantial efficiency gain. It can be attributed to the efficient retrieval architecture in MobileRAG, where LocalRAG requires only \textbf{0.005} seconds per retrieval and interRAG only \textbf{0.4} seconds, thereby significantly decreasing the number of LLM-based agent calls and accelerating the overall runtime. With the support of LocalRAG, MobileRAG significantly reduces the number of steps to open apps, especially for multi-app tasks. By enabling direct retrieval and execution of local apps, LocalRAG greatly enhances the efficiency of MobileRAG. Additionally, fewer operational steps reduce the likelihood of operational errors or omissions, thereby enhancing the overall architectural performance.

\begin{table}[htbp]
  \setlength{\tabcolsep}{1 mm}
		\begin{tabular}{l|ccccccc}
			\toprule
			Model &   AS & AF & RP & TCR & TSR & avg & open app\\ 
			\midrule
			w/o Local    & 93.8 & 75.4& 96.8 & 82.1 & 67.5 & 11.42 & 3.12\\
                w/o Inter    & 88.1 &  85.3 & 97.1 &79.7
                & 65.0 & 9.94 & 1.81\\
			\midrule
			MobileRAG  &100 & 86.4 & 98.5 & 91.2 &80.0&9.15 & 1.86\\
			\bottomrule     
		\end{tabular}
		\caption{Ablation studies of LocalRAG and InterRAG. ``w/o Local'' and ``w/o Inter''  indicate the removal of LocalRAG and InterRAG, respectively.}
		\label{abla}	
\end{table}

\textbf{Ablation Studies.} Table \ref{abla} shows the results of Ablation studies, we find each component of MobileRAG plays a key role in improving performance. First, we use LLM-based visual perception instead of LocalRAG. The results show varying degrees of decline across different metrics, demonstrating that LocalRAG achieves significant performance improvements compared to  GUI-based mobile agent, particularly in terms of interaction steps. LocalRAG can directly validate and operate locally installed apps, thereby eliminating unnecessary steps such as manual searching and interface understanding. Second, we remove InterRAG in our framework, we observed a decrease of 11.9\% and 18.75\% in AS and TSR, respectively. This finding confirms that InterRAG is capable of accessing the updated and comprehensive app information from external sources, thereby enabling more informed and accurate app selection.

We further analyzed the effectiveness of MemRAG by testing 15 groups of data, encompassing cases with complete consistency, similarity above 80\%, and similarity below 80\% (More detail are provided in Appendix \textcolor{red}{B}.). As shown in Table \ref{abla2}, MemRAG effectively enhances model performance, particularly by reducing the number of operational steps, with an average reduction of 2.4 steps. These results clearly demonstrate that MobileRAG is capable of learning relevant operational procedures from historical successful steps via MemRAG, thereby improving both efficiency and overall task accuracy.

\begin{table}[htbp!]

		\begin{tabular}{l|cccccc}
			\toprule
			Model &   AS & AF & RP & TCR & TSR & Step\\ 
			\midrule
			w/o Mem      & 100 & 88.7 & 98.4 & 93.8 & 86.7 & 10.6\\
			with Mem    & 100 & 91.1 & 98.7 & 97.6 &  93.3 & 8.2\\
    \bottomrule 
		\end{tabular}

		\caption{Ablation studies of MemRAG. ``w/o Mem'' indicates the removal of MemRAG.}
		\label{abla2}	
\end{table}

\textbf{Case Study.} Figure \ref{exp1} demonstrates the reasoning process of MobileRAG when no suitable match is found by MemRAG. We find that MobileRAG has the following advantages: 1) utilizing InterRAG enables comprehensive understanding of user queries;  2) leveraging LocalRAG allows the system to directly launch relevant applications from the local app library, minimizing accidental touches or erroneous actions; 3) MobileRAG’s robust retrieval capabilities help reduce the number of operational steps, particularly benefiting long sequential tasks.  Figure \ref{case_mem} illustrates the inference process when MemRAG successfully identifies a matching instance. Through MemRAG, It can effectively extract and leverage information from historically correct operation sequences, thereby enhancing agent performance and further decreasing the number of operational steps.

\section{Conclusion}

In this paper, we introduce MobileRAG, a novel framework that enhances mobile agents with Retrieval-Augmented Generation (RAG). MobileRAG is designed to address several limitations of current LLM-based mobile agents, such as over-reliance on language model comprehension, lack of interaction with external environments, and absence of effective memory. It integrates three components—InterRAG, LocalRAG, and MemRAG—to enable more accurate understanding of user queries and efficient execution of complex mobile tasks. Additionally, we propose MobileRAG-Eval, a challenging benchmark featuring real-world mobile scenarios that require external knowledge support. Extensive experiments on MobileRAG-Eval demonstrate that MobileRAG outperforms state-of-the-art methods by 10.3\%, achieving better results with fewer operational steps.

\bibliography{aaai2026}
\newpage
\section{Appendix A: Benchmark Details}

Our comprehensive benchmark evaluates mobile agents across 80 distinct task instructions. We designed and introduced 36 novel task instructions (outlined in Table \ref{tab:Original_task_instructions}) specifically crafted for this benchmark, complemented by 44 supplementary instructions sourced from Mobile-Agent-v2 that address standard operations. Together, these instructions form a robust evaluation framework that comprehensively assesses mobile device operations across a wide range of scenarios.

\begin{table*}[!h]
  \centering
  \scalebox{0.85}{
    \begin{tabular}{
      >{\centering\arraybackslash}m{8.5em}   
      >{\arraybackslash}m{24em}    
      >{\arraybackslash}m{24em}}   
    \toprule
    \multicolumn{1}{c}{\textbf{Category}} & \multicolumn{1}{c}{\textbf{Basic Instructions}} & \multicolumn{1}{c}{\textbf{Advanced Instructions}} \\
    \midrule
    Common & 1. Set an alarm for 8 am. \newline{}2. Write a reminder: I have dinner tonight at 6:30. \newline{}3. Find a short, trendy video tutorial on cooking steak and like it. & 1. Find a coffee shop near me on Google maps, and then open Google message to tell Mike that I will be at the place waiting for him (give him the specific coffee shop name).\newline{}2. Summarize the prices of three nearby gas stations in `Google Maps' app and record these information into the `Notepad' app. \newline{}3. Find me a coffee shop near me on 'Google Maps' app that sells birthday cakes and is within a 10-minute drive. Find the phone number and create a new note for it in the `Notes' app. \\
    \midrule
    Music\_single & 1. Listen to a piano music in the`YouTube Music' APP.\newline{}2. Play exclusive album `Ear'. \newline{}3. Play `Shape of You' by Ed Sheeran on Spotify. & 1. Find the album `Happier Than Ever', and add it to my library.\newline{}2. Play Taylor Swift's `Love Story', and add this song to a new playlist named `Agent Playlist'.\newline{}3. Add the Taylor Swift's song ``1989 (Taylor's Version)" to playlist \\
    \midrule
    Movie\_single & 1.Like the TV series `Good Boy' season 1 in the `Prime Video' app.\newline{}2. Add the TV series `The Boys' to the viewing list.\newline{}3. Add a comedy movie to the Watchlist in the `Prime Video' app. & 1. Add the movie `Heads of state.' to the Watchlist and like it\newline{}2. Watch the TV series `A Dream Within A Dream' and comments it. \newline{}3. Find the TV series `The Boys', select season 3 and add it to the Watchlist. \\
    \midrule
    Exteral\_information & 1. What is the NBA Score today? Send the results to Jelly. \newline{}2. Send a message to Jelly to tell her the app that can watch Squid Game.\newline{}3. What is the USD to CNY exchange rate? Find the specific price rather than link, send a message to Jelly to tell her the results. & 1. Download the app to watch Squid Game.\newline{}2. Download the most popular music app and open it. \newline{}3. Search for the date of the next Winter Olympics opening ceremony and then set a reminder for that date in the 'Calendar' app. \\
    \midrule
    Movie\_multi & 1. Open the `Prime Video' app to find the movie in my Watchlist, and then search for this movie in X, choose the first posts to read and comment it, then send a message to invite Jelly to watch this movie in 'Google Message' app.\newline{}2. Find the TV series `The Boys', select season 3 and add it to the Watchlist, then summarize the introduction, and send a message to Mike inviting him to watch `The Boys' season 3, tell him the introduction.\newline{}3. Please open YouTube to search for one short video about `Heads of State' and like it, then find an app that can watch the movie `Heads of State', and add it to the Watchilist.  & 1. Open the app that can watch `A Dream Within A Dream', then find the top 3 comments, write these comments into the notepad.\newline{}2. I need to find an app to watch the TV series `The Boys', then find its introduction, and finally create a note to summarize the introduction.\newline{}3. Open the app that can watch `A Dream Within A Dream' in the `IQIYI' app, then find details like director and rating score, then message to Jelly the info through 'Google Message' app. \\
    \midrule
    Music\_multi & 1. Play a piano music in the `YouTube Music' app and then open the `X' app to search for the song name.\newline{}2. Open the `Spotify' app to find the song 'Shape of You' by Ed Sheeran and add it to liked song, then share this song to Mike and invite him to join on Spotify through 'Google Message' app.\newline{}3. Add Taylor Swift's song  ``1989 (Taylor's Version)" to playlist, and then search for this song in the `X' app, follow one related account and enter his/her post. & 1. Open two local music apps that can play songs. Check if the song ``Happier Than Ever" is available in each of these apps. Then, record the availability status of the song in a notepad app.\newline{}2. Open two local music apps that can play songs. Check if Taylor Swift's song ``1989 (Taylor's Version)" is available in each of these apps, if it is add this song into playlist. Then, send a message to Mike to tell him the availability status of the song on Google Message.\newline{}3. Open three local music apps that can play songs. Check if the song ``Happier Than Ever" is available in each of these apps. Then, record the availability status of the song in a notepad app. \\
    \bottomrule
    \end{tabular}%
        }
  \caption{The 36 Original Task Instructions Specifically Designed for Our Mobile Agent Benchmark}
  \label{tab:Original_task_instructions}%
\end{table*}%

\section{Appendix B: MemRAG Effectiveness Analysis}
As shown in Table \ref{tab:memRAG_task}, we conducted a detailed evaluation of MemRAG's effectiveness using 15 distinct task groups, systematically categorized into three similarity levels:
\begin{itemize}
    \item \textbf{Original Instructions}: Tasks executed in historical operations
    \item \textbf{High Similarity (exceeding 80\%)}: Tasks sharing core operational patterns with historical cases
    \item \textbf{Low Similarity (below 80\%)}: Tasks with varying degrees of novelty
\end{itemize}
The task design systematically varies complexity while maintaining real-world applicability, allowing for a detailed evaluation of MemRAG's retrieval-augmented operational capabilities across the spectrum of task similarity.
\begin{table*}[htbp]
  \centering
    \begin{tabular}{
      >{\arraybackslash}m{16.25em}   
      >{\arraybackslash}m{16.25em}    
      >{\arraybackslash}m{16.25em}}   
    \toprule
    \multicolumn{1}{c}{\textbf{Original Instructions}} & \multicolumn{1}{c}{\textbf{80-100\% Similarity Instructions}} & \multicolumn{1}{c}{\textbf{0-80\% Similarity Instructions}} \\
    \midrule
    Send a greeting message to Jelly. & Send a greeting message to Jelly. & Play `Shape of You' by Ed Sheeran on Spotify. \\
    \midrule
    Like the TV series `Good Boy' season 1 in the `prime video' app. & Send a message to Jelly to let her know which app can be used to watch Squid Game. & Search for the date of the next Winter Olympics opening ceremony and then set a reminder for that date in the 'Calendar' app. \\
    \midrule
    Download the app to watch Squid Game. & Add a comedy movie to the Watchlist in the `prime video' app. & Summarize the prices of three nearby gas stations in `Google maps' app and record these information into the in the 'Notepad' app . \\
    \midrule
    Find the TV series `The Boys', select season 3 and add it to the Watchlist, then summarize the introduction, and send a message to Mike inviting him to watch `The Boys' season 3, tell him the introduction. & Send a message to Jelly to invite him to watch `The Boys' season 3, tell him the introduction. & Please open YouTube to search for one short video about `Heads of State' and like it, and then find an app that can watch movie `Heads of State', and add it to the Watchilist. \\
    \midrule
    Open two local music apps that can play songs. Check if the song `Happier Than Ever' is available in each of these apps. Then, record the availability status of the song in the `Notepad' app. & Open three local music apps that can play songs. Check if the song 'Happier Than Ever' is available in each of these apps. Then, record the availability status of the song in in the `Notepad' app. & Play a piano music in the `YouTube Music' app and then open the `X' app to search for the song name. \\
    \bottomrule
    \end{tabular}%
  \caption{MemRAG Task Instruction}
  \label{tab:memRAG_task}%
\end{table*}%

\section{Appendix C: Case Study and Test Results}

Appendix C presents a detailed case study showcasing several test results to demonstrate the performance of MobileRAG. 

\begin{figure*}[!h]
    \centering
    \includegraphics[width=0.9\linewidth]{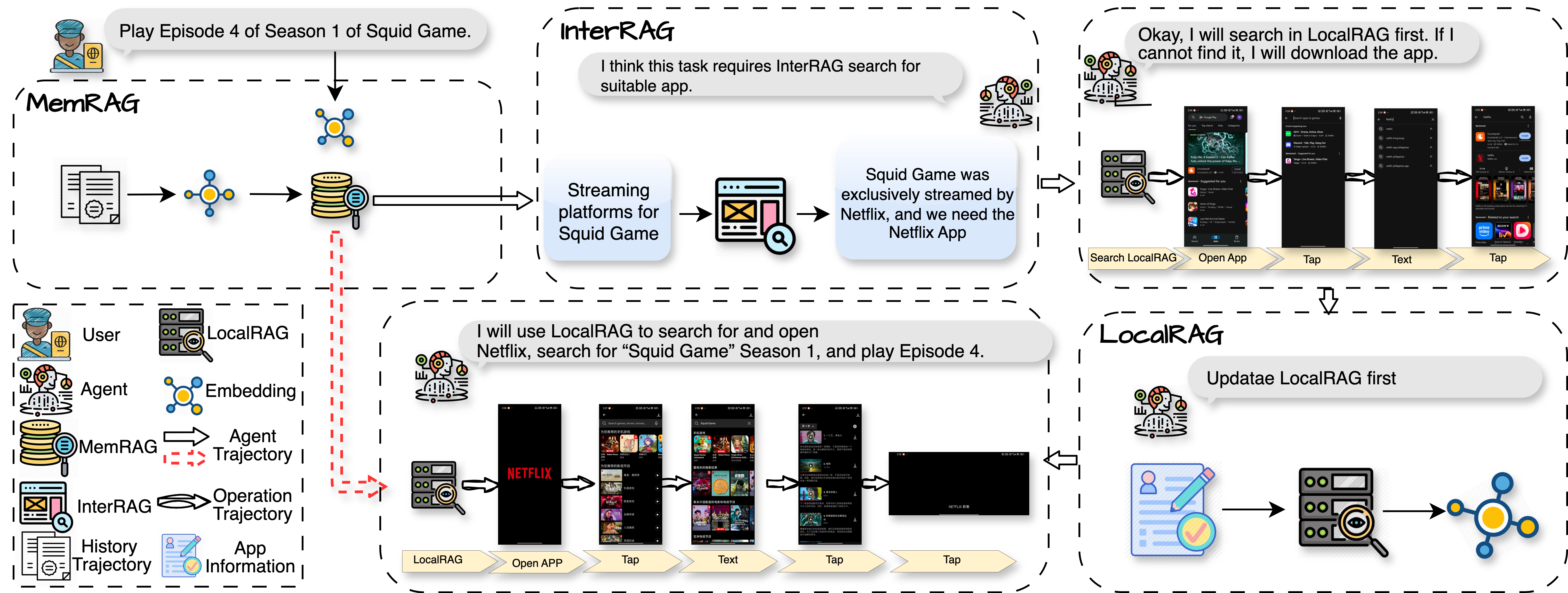}
    \caption{Framework Diagram on the Real Machine - Xiaomi 15 Pro.}
    \label{Appendix_framework}
\end{figure*}

Figure \ref{Appendix_framework} shows the overall framework of MobileRAG is tested on a real Xiaomi 15 Pro device. This diagram provides an overview of how MobileRAG operates, emphasizing its architecture and functionality in real-time testing.

\begin{figure*}[!htbp]
    \centering
    \includegraphics[width=0.9\linewidth]{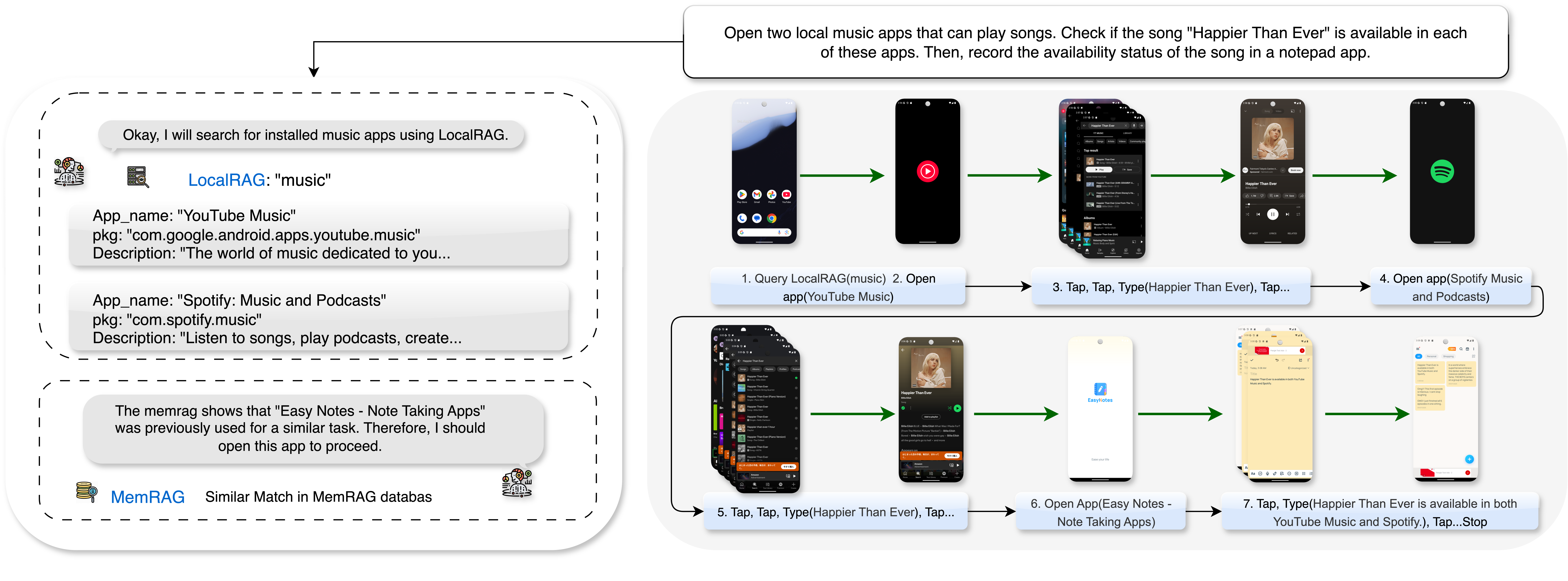}
    \caption{This case study explores the workflow integration across three distinct mobile applications.}
    \label{Appendix_3app_task}
\end{figure*}

Figure \ref{Appendix_3app_task} showcases the testing process with three different apps, highlighting the superiority of LocalRAG in app retrieval. The performance improvements and efficiency of LocalRAG are clearly depicted, focusing on its ability to retrieve app data effectively.

\begin{figure*}[!htbp]
    \centering
    \includegraphics[width=0.9\linewidth]{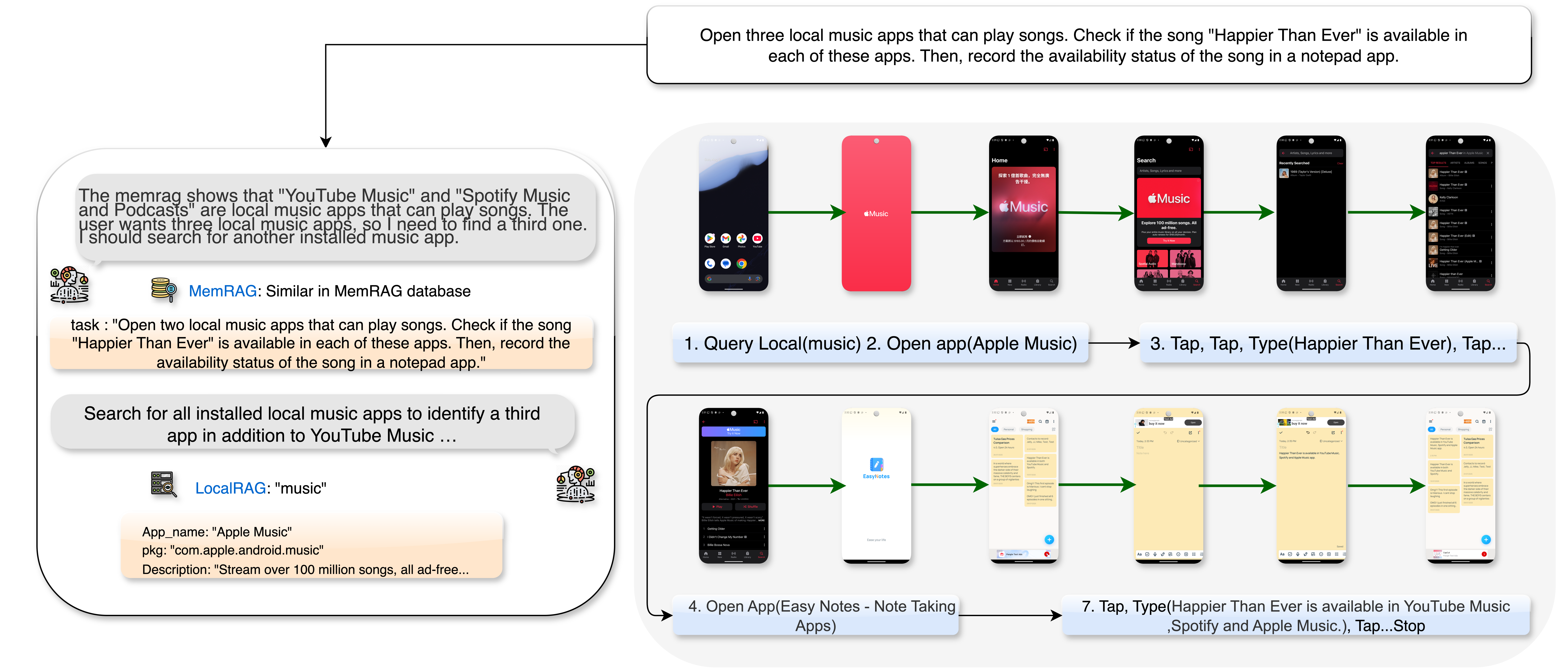}
    \caption{The diagram represents the workflow of multiple apps operating simultaneously with MemRAG support.}
    \label{Appendix_3app_mem}
\end{figure*}

Figure \ref{Appendix_3app_mem} emphasizes how MemRAG enhances the retrieval of past successful experiences building upon the previous figure’s instructions, significantly reducing the steps and complexity involved in operations. The graph illustrates how MemRAG streamlines the process by reusing previously successful tasks, improving the user experience by minimizing effort and time.

The integrated RAG architecture offers synergistic advantages: InterRAG addresses ambiguous instructions by incorporating external context to ensure precise app selection. LocalRAG facilitates accurate app targeting, while MemRAG leverages historical successes to retrieve optimal operational sequences, thereby reducing both cognitive load and execution complexity.

\end{document}